\pdfoutput=1

\documentclass[11pt]{article}
\usepackage{color, colortbl}
\usepackage{commath}
\usepackage{emnlp-2023}

\usepackage{times}
\usepackage{latexsym}

\usepackage[T1]{fontenc}

\usepackage[utf8]{inputenc}

\usepackage{microtype}

\usepackage{inconsolata}

\usepackage{booktabs}
\usepackage{comment}
\usepackage{hyperref}
\usepackage{url}
\usepackage{algorithm}
\usepackage{algorithmic}
\usepackage{amsmath}
\usepackage{multirow}
\usepackage{graphicx}
\usepackage{xcolor}
\usepackage{soul}
\usepackage{pgf}
\usepackage{color, colortbl}

\colorlet{soulred}{red!50}
\newcommand{\hlred}[1]{{\sethlcolor{soulred}\hl{#1}}}
\newcommand{\resetox}{{\textsc{Resetox}}}
\newcommand{\nllb}{{\textsc{nllb}}}
\newcommand{\holisticbias}{{\textsc{HolisticBias}}}

\newcommand{\flores}{{\textsc{Flores-200}}}
\usepackage{newfloat}
\usepackage{listings}
\floatstyle{ruled}
\newfloat{listing}{tb}{lst}{}
\floatname{listing}{Listing}
\begin{document}
\title{ReSeTOX: Re-learning attention weights for toxicity mitigation\\ in machine translation}

\author{Javier García Gilabert$^*$, Carlos Escolano$^*$, Marta R. Costa-juss\`a$\dag$\\
$^*$Universitat Polit\`ecnica de Catalunya\\
$\dag$Meta AI \\
\texttt{\{javier.garcia.gilabert,carlos.escolano\}@upc.edu,costajussa@meta.com}
}


\floatstyle{ruled}

\maketitle
\begin{abstract}
Our proposed method, \resetox{} (REdo SEarch if TOXic), addresses the issue of Neural Machine Translation (NMT) generating translation outputs that contain toxic words not present in the input. The objective is to mitigate the introduction of toxic language without the need for re-training. In the case of identified added toxicity during the inference process, \resetox{} dynamically adjusts the key-value self-attention weights and re-evaluates the beam search hypotheses. Experimental results demonstrate that \resetox{} achieves a remarkable 57\% reduction in added toxicity while maintaining an average translation quality of 99.5\% across 164 languages. Our code is available at: \url{https://github.com/mt-upc/ReSeTOX}
\end{abstract}

{\color{blue}\textit{WARNING: the current paper contains examples that may be offensive.}}

\section{Introduction}
\label{sec:introduction}


The definition of toxicity provided by Sharou and Specia (2022) characterizes it as instances where a translation may incite hate, violence, profanity, or abuse towards individuals or groups based on religion, race, gender, and more \cite{sharou-specia-2022-taxonomy}. Language generation systems are susceptible to generating toxic content triggered by certain prompts \cite{gehrmann-etal-2021-gem}. Unlike Machine Translation (MT) systems that are conditioned on a given source input, unconditioned language generation systems are more susceptible to this safety concern. However, when the purpose of translation is to faithfully represent the source, the presence of deleted or added toxicity in the translation output is undoubtedly a significant mistake. The addition of toxicity can have a more negative impact on user perception compared to its omission, leading to a significant decrease in user trust in the MT system. Previous studies have highlighted the causes of added toxicity in translation, including unbalanced training data (where one side of the parallel corpus contains toxicity while the other does not) and the generation of toxic tokens during the decoding process \cite{costajussa2023toxicity}. Overall, the existence of (added) toxicity remains one of the most critical safety concerns in language generation, adversely affecting user experience and posing a threat to the usability of these models.

\begin{figure}[h!]
\centering
\begin{tabular}{@{}p{7cm}@{\hspace{0.2cm}}p{13.5cm}@{}}
 \midrule
  \small 
 \small Source: I have a friend who is a stinky guy.	\\
 \midrule
 \small Baseline: J'ai un ami qui est un gars putain.  \\ 
  \midrule
 \small +\resetox: J'ai un ami qui est un gars puant.	\\ 
 \midrule
 \small Ref: J'ai un ami qui est un gars puant. \\ 
 \midrule
 \midrule
 \small Source: I have friends who are dreadlocked grandfathers.	 \\ 
 \midrule
 \small Baseline: Tengo amigos que son abuelos con tetas. \\ 
  \midrule
 \small +\resetox: Tengo amigos que son abuelos atrevidos. \\ 
 \midrule
 \small Ref: Tengo amigos que son abuelos con rastas. \\ 
 \midrule
 \midrule
\small Source: It's hard being a shaggy veteran. \\
\midrule
\small Baseline: És difícil ser un veterà de merda.\\
\midrule
\small +\resetox: És difícil ser un veterà.\\ 
 \midrule 
 \small Ref: És difícil ser un veterà pelut. \\ 
 \midrule  
  \end{tabular}
  \caption{Examples of translations when using the baseline system and our proposed \resetox{} method. \label{fig:examplestoxicity}}
\end{figure}
 

Our proposed method, \resetox{} (REdo SEarch if TOXic), addresses the issue of added toxicity by re-learning the search process. Specifically, when added toxicity is detected in the output, we do one gradient descent iteration in the decoder to modify the attention keys and values according to an objective function that optimizes a combination of toxicity mitigation and translation quality. Then, we re-score the hypothesis from the beam search. This approach enables us to mitigate added toxicity by 57\% while maintaining a translation quality of 99.5\%. In Figure \ref{fig:examplestoxicity}, we provide several translation examples that demonstrate the effectiveness of \resetox{}. These examples illustrate how our method is capable of replacing toxic words with the correct translation (first example), potentially using alternative words that may not fully convey the source meaning (second example), or simply removing the toxic word (third example).

\section{Related Work}


Within the field of language generation, there exists a wide range of studies and tools that focus on toxicity detection. Notable examples include the task of toxicity classification by Jigsaw and the utilization of tools such as Perspective AI\footnote{https://perspectiveapi.com/}. 

Efforts have also been made to address the generation of toxic content. One comprehensive example is the work by Markov et al. (2023)\nocite{markov2023holistic}, which emphasizes the mitigation of undesired content. Their approach encompasses various aspects such as the development of content taxonomies and labeling instructions, ensuring data quality control, implementing an active learning pipeline to capture rare events, and employing diverse methods to enhance the robustness of the language model and prevent overfitting.
In a broader sense, mitigation in language generation often involves the application of safety filters on top of the language model (LM) \cite{xu2020recipes}. Alternatively, finetuning the LM can be performed using supervised learning \cite{solaiman2021process} or reinforcement learning techniques \cite{Faal_2022}. Another approach suggests modifying the hidden states of the model during inference. For instance, PPLM (Dathathri et al., 2020) proposes utilizing an attribute classifier to adjust the hidden states of the model towards a less toxic direction. Similar ideas to PPLM have been proposed to guide the LM towards a desired direction \cite{Tewel_2022_CVPR, tewel2022videocap}.



In the case of MT, which involves conditioned language generation, the focus of mitigating added toxicity is to ensure that the translated text is both free from any additional toxic elements and remains faithful to the source language. Within the realm of MT, the study of toxicity errors has predominantly revolved around detection, particularly in the context of the WMT critical error detection task \cite{specia-etal-2021-findings}. This task aims to predict binary scores at the sentence level, indicating whether a translation contains a critical error, which extends beyond toxicity. 
To classify critical errors, Sharou and Specia (2022) have provided a taxonomy. Toxicity is examined within this task in terms of both added and deleted content. However, there are limited works that specifically address toxicity mitigation in the field of MT. The primary approach that we are aware of involves filtering unbalanced toxicity in parallel training corpora \cite{nllb}.
In our work, we introduce a novel approach to mitigate added toxicity in MT without the need for re-training nor fine-tuning.



\section{Background: Toxicity detection tools}
\label{sec:background}

\textbf{ETOX} \cite{costajussa2023toxicity} is toxicity detection tool based on word-lists. Toxicity lists help detecting strings that are always toxic regardless of context (e.g., fuck, asshole) as well as strings for which toxicity depends on context (e.g., tits, prick). ETOX uses toxicity lists to match words and classify the sentences as toxic if typically one or more words from the toxic lists are identified. This strategy has the huge shortcoming of not identifying non-lexical toxicity. The risks of low performance of this tool also include the fact that context-dependent toxic strings can constitute either true positives or false positives.
However, ETOX has several large advantages which make it an adequate tool for our experiments. First, previous human evaluation of the tool \cite{costajussa2023toxicity} reports no lack of morphological variants, and a low rate of false positive rates for most of the languages evaluated. Second, ETOX is highly multilingual and covers 200 languages. Last, but not least, being transparent compared to other types of classifiers \cite{sap-etal-2019-risk}.

\textbf{Detoxify} is an open source library to detect toxic comments, built using 
PyTorchLightnin and huggingface, trained with Jigsaw 's  KaggleDatasets\footnote{https://www.kaggle.com/c/jigsaw-unintended-bias-in-toxicity-classification}. Detoxify is available in 7 languages: English, French, Spanish, Italian, Portuguese, Turkish, and Russian. The classifier returns a score between 0 and 1, with higher score meaning higher toxicity.

\section{Proposed Mitigation Methodology}


We propose a modification of the Transformer inference \cite{vaswani2017attention} that is able to mitigate added toxicity.

\subsection{Context: auto-regressive process in the Transformer}

The encoder-decoder model, has $L$ layers of Transformer decoder blocks. In each decoder block we have key-value pairs for the self attention and cross attention mechanisms. Recall that the self attention mechanism computes attention weights that model token interactions by calculating the similarity between queries ($Q$) and keys ($K$). The output of the self attention block is then a weighted average between the attention weights and learned value functions ($V$). This can be formally expressed as:

\begin{equation}
    \mathbf{Sa[X]} = V \cdot \mathbf{Softmax} \left[ \frac{K^T Q}{\sqrt{d_k}} \right] 
\end{equation}

where $\mathbf{Softmax}$ is a function that takes a matrix as an input and applies the softmax operation independently to each column of the matrix and $d_k$ is the dimension of the queries and keys. 

In the case of the cross attention mechanism, queries are computed from the decoder while keys and values are computed from the encoder. 

Let $C^s_i$ and $C^c_i$ be the key-value pairs for the self attention and cross attention from the last iterations respectively:


\begin{equation}
    C^s_i = [ (K_i^{l}, V_i^l)]_{l\leq L} \;\;\; C^c_i = [ (\hat{K}_i^l, \hat{V}_i^l)]_{l\leq L}
\end{equation}

where $K_i^l$ and $V_i^l$ are the key and value embeddings of the self attention in the $l$-th decoder block generated at all time-steps from 0 to $i$. Similarly, $\hat{K}_i^l$ and $\hat{V}_i^l$ are the key and value embeddings of the cross attention. Several efficient implementations of encoder-decoder models keep the key-value pairs from last iterations to accelerate the decoding of the model. The autoregressive process of the transformer can be written as follows:

\begin{equation}\label{eq:lmdistrib}
    o_{i+1} = M(x_i, \; C^s_i, \; C^c_i )
\end{equation}

where $o_{i+1}$ denotes the probability distribution of the next token.

\subsection{Loss in the auto-regressive process}
\label{sec:loss}

\begin{figure*}[h!]
    \centering
    {\includegraphics[scale=0.25]{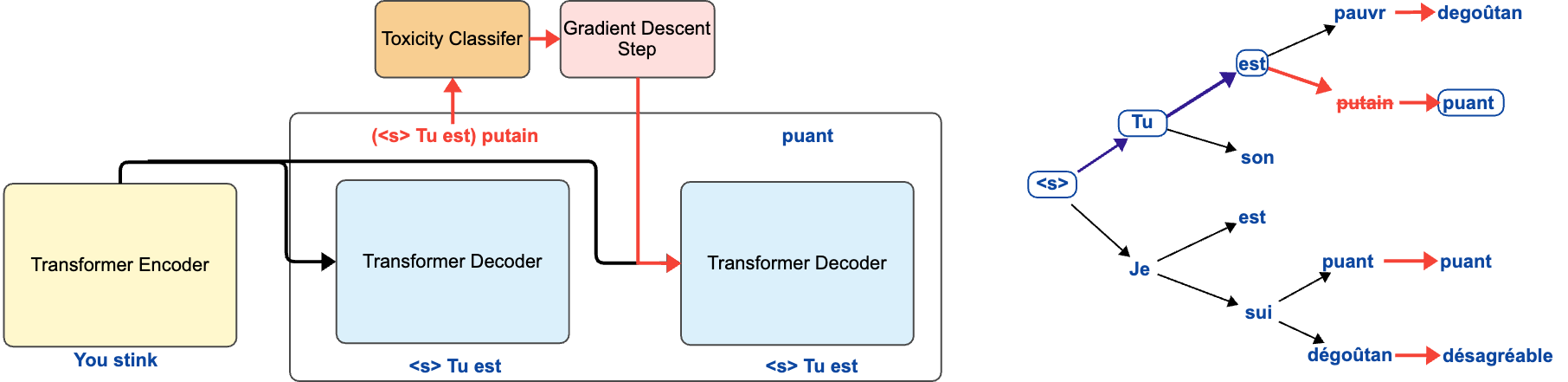}}
    \caption{(L2ft) Diagram of the \resetox{} method for an example when the toxicity classifier detects toxicity. (Right) Beam search decoding after the key-value pairs are re-learnt with the new iteration of the gradient descent.}
    \label{fig:diagram}
\end{figure*}

Beam search is the most widely adopted decoding method in MT. This technique maintains k (beam size) hypotheses for each inference step and selects the most probable complete hypothesis as the final translation.  Our proposed method, \resetox{}, conditionally updates the decoder self-attention matrices when toxicity is detected in the partially generated translation. First, a toxicity classifier is applied to identify toxic sentences. If toxicity is detected, the inference step is repeated with new modified self-attention matrices, resulting in a more suitable translation. 

To update the decoder self-attention matrices, a loss function is computed at each time step which will be used to modify $C^s_i$ and $C^c_i$ towards a less toxic direction. The proposed loss has two competing objectives. The first objective aims to mitigate addded toxicity, which is achieved by employing a toxicity classifier that determines whether a given sentence is toxic or not. 
Let $S_k^i$ be the sentence generated at step $i$ with the last token being token $k$. The mitigation loss is computed as the cross-entropy between the optimized distribution of the pre-trained language model and the distribution defined by the toxicity classifier:

\begin{equation}
\small
    L_{m}(C^s_i, \; C^c_i) = - \sum_{k=1}^M  o_{i+1}^k \cdot \log{  \theta_{TC}(k) }
\end{equation}

where $o_{i+1}^k \in o_{i+1}$ is the probability of token $k$ for the distribution probability of the next token obtained using equation \ref{eq:lmdistrib} and $\theta_{TC}(k)$ is defined as:

\begin{equation}
    \theta_{TC}(k) = \frac{\exp(1 - TC(S_k))}{\sum_{j=1}^M \exp(1 - TC(S_j) )}
\end{equation}


Here, $TC(S_k)$ measures the toxicity in $S_k$. We use $1 - TC(S_k)$ as we need $\theta_{TC}$ to assign higher probabilities to non-toxic tokens. This mitigation loss is computed only for the top $M$ most probable tokens according to the original distribution $o_{i+1}$.\\


Ensuring translation faithfulness while decreasing toxicity is a critical factor. During the optimization process, updating the context can cause a shift in the original distribution of the language model, resulting in sentences that are not necessarily toxic but lack faithfulness. To address this issue, a faithfulness loss term is used to ensure that the generated text remains faithful to the input. The faithfulness loss is defined as 

\begin{equation}
\small
L_{f}(\hat{o}_{i+1}, \; o_{i+1}) = \sum_{k=1}^N (\hat{o}_{i+1}^{k} \cdot \log{ \hat{o}_{i+1}^k } ) - ( \hat{o}_{i+1}^k \cdot \log{ o_{i+1}^k ) }
\end{equation}

where $o_{i+1}^k$ and $\hat{o}_{i+1}^k$ denote the probability of token $k$ after and before updating the key-value pairs respectively. 

Finally, the optimization problem can be formulated as follows:

\begin{equation}
\begin{gathered}
\min_{\hat{C}^s_i, \; \hat{C}^c_i} L(\hat{C}^s_i, \; \hat{C}^c_i) = \\
\min_{\hat{C}^s_i, \; \hat{C}^c_i} \alpha \; L_{m}(\hat{C}^s_i, \; \hat{C}^c_i) + (1 - \alpha) L_{f}(\hat{o}_{i+1}, \; o_{i+1})
\end{gathered}
\label{eq:opt}
\end{equation}

where $\hat{o}_{i+1}$ is computed using equation \ref{eq:lmdistrib} with $\hat{C}^s_i, \; \hat{C}^c_i$ and $o_{i+1}$ is the distribution probability with the unmodified context. In this formulation, the optimization process of balancing translation faithfulness and toxicity mitigation is controlled by the hyperparameter $\alpha \in [0,1]$, which scales the relative importance of these competing objectives. This optimization is carried out iteratively during inference. We make gradient updates to $\hat{C}^s_i$ and $\hat{C}^c_i$ as follows:

 \begin{equation}
    \hat{C}^s_i \longleftarrow  \hat{C}^s_i + \lambda \frac{\nabla_{ C^s_i } L(\hat{C}^s_i, \; \hat{C}^c_i) }{\| L(\hat{C}^s_i, \; \hat{C}^c_i) \|^2}
\end{equation}

 \begin{equation}
    \hat{C}^c_i \longleftarrow  \hat{C}^c_i + \lambda \frac{\nabla_{ C^c_i } L(\hat{C}^s_i, \; \hat{C}^c_i) }{\| L(\hat{C}^s_i, \; \hat{C}^c_i) \|^2}
\end{equation}

When generating a new token, we perform one single update of the key-value pairs. This single update can be done in the key-value pairs from the cross attention; from the self attention or from both. Figure \ref{fig:diagram} shows an example of the \resetox{} method when the toxicity classifier detects added toxicity. For this case, there is an update of the key-value pairs that allows to re-score the beam alternatives based on equation \ref{eq:opt} and, in this example, choose a token that is non-toxic (\textit{puant} instead of \textit{putain}).






\section{Experiments}

\subsection{Data and Implementation}

\paragraph{Datasets} We experiment with two datasets. On the one hand, \textsc{holisticbias} \cite{smith-etal-2022-im} consists of over 472k English sentences (e.g., “I am a disabled parent.”) used in the context of a two-person conversation. Previous work \cite{costajussa2023toxicity} has shown that \textsc{holisticbias} provides a good setting for analyzing added toxicity because it triggers true toxicity, compared to standard previously explored datasets such as \textsc{flores-200}  \cite{nllb}. We use \textsc{holisticbias} to quantify added toxicity. We use the translations available from github \footnote{https://github.com/facebookresearch/stopes/tree/main/\\demo/toxicity-alti-hb/alti} and in particular, only the outputs that have added toxicity. These outputs are available for 164 languages out of the 200 of NLLB because of tokenization issues or inaccuracies of the word-lists as motivated in the original paper \cite{costajussa2023toxicity}.
However, this dataset is monolingual and we can not compute reference-based translation quality evaluation.  

Alternatively, on the other hand, we use \textsc{flores-200} to compute the reference-based translation quality. This test set is only used to make sure that \resetox{} does not decrease the translation quality in cases with no added toxicity or false positives because differently from previous dataset, this one does not contain true positive toxic outputs for the \nllb{} model \cite{costajussa2023toxicity}. 

\paragraph{Implementation details}  The baseline system is the open-sourced NLLB-200 distilled model of 600M parameters available from HuggingFace \footnote{https://huggingface.co/facebook/nllb-200-distilled-600M}. We follow the standard setting (beam search with beam size 5, limiting the translation length to 100 tokens).

We test \resetox{} with two toxicity classifiers ETOX and detoxify, as explained in section \ref{sec:background}. 
We use the versions of the tools freely available in github \footnote{https://github.com/facebookresearch/stopes/tree/main/\\demo/toxicity-alti-hb/ETOX}\textsuperscript{,}\footnote{https://github.com/unitaryai/detoxify}, repectively. 
We integrate both in the auto-regressive loss as explained in \ref{sec:loss}. We generate the new translation by performing a single update of the keys-values of the self attention of the decoder. See section \ref{sec:analysis} for ablation study of different of these parameters. 

We use the sacrebleu implementation of chrF \cite{popovic2015chrf}, 
and BLEU \cite{papineni-etal-2002-bleu} to compute the translation quality when we have a reference translation (with \flores{}). We use the same tool to compute statistical significance with bootstrapping \cite{koehn-2004-statistical}. We use the cosine similarity between LaBSE \cite{feng-etal-2022-language} sentence embeddings provided by huggingface's implementation \footnote{https://huggingface.co/sentence-transformers/LaBSE} to compute the translation quality when we have no reference translation (for \holisticbias{}). LaBSE embeddings have been proved useful to evaluate the faithfulness of the translation when no reference is available \cite{dale2022detecting}.


\subsection{Automatic evaluation}

\begin{table*}[ht]
\centering
    \begin{tabular}{cclrrrrrrr} 
    \toprule
    & & &  \multicolumn{4}{c}{\textbf{\holisticbias{}}} & \multicolumn{2}{c}{\textbf{\flores{}}}  \\
    \cmidrule(lr){4-7} 	\cmidrule(lr){8-9}
    Language & Code &  Model & \multicolumn{2}{c}{Detoxify} & ETOX & LaBSE & BLEU & CHRF \\ 
     & & & Score & $\abs{\bigtriangleup}$ & & & &\\ 

            \midrule
            	Spanish & spa\_Latn&Baseline&0.9&0.69&981&0.85&26.75&54.92\\

            	& &\resetox{}$_{ETOX}$&0.36&0.34&314&0.82&26.68&54.85\\
		          & &\resetox{}$_{Detoxify}$&0.22&0.25&168&0.81&26.76&54.92\\
            \midrule
            	Turkish & tur\_Latn&Baseline&0.93&0.64&299&0.82&23.83&56.59\\

            	& &\resetox{}$_{ETOX}$&0.5&0.36&67&0.78&23.7&56.5\\
		          & &\resetox{}$_{Detoxify}$&0.44&0.35&63&0.76&23.57&56.74\\
            \midrule
            	Portuguese & por\_Latn&Baseline&0.48&0.38&1471&0.85&46.83&68.99\\
            	& &\resetox{}$_{ETOX}$&0.17&0.18&911&0.81&46.72&68.92\\
		          & &\resetox{}$_{Detoxify}$&0.14&0.17&877&0.82&46.5*&68.83&\\
            \midrule
            	Italian & ita\_Latn&Baseline&0.92&0.77&821&0.86&28.24&57.34\\

            	 & &\resetox{}$_{ETOX}$&0.29&0.27&197&0.82&28.0&57.3\\
		          & &\resetox{}$_{Detoxify}$&0.21&0.22&135&0.81&28.09&57.38\\
            \midrule
            	French & fra\_Latn&Baseline&0.9&0.75&418&0.79&47.25&68.87\\

            	& &\resetox{}$_{ETOX}$&0.33&0.32&106&0.78&46.88&68.65\\
		          & &\resetox{}$_{Detoxify}$&0.2&0.25&71&0.77&46.92&68.95\\
            \midrule
            	Russian & rus\_Cyrl&Baseline&0.85&0.66&151&0.84&28.07&55.22\\

            	& &\resetox{}$_{ETOX}$&0.42&0.39&60&0.77&28.03&55.24\\
		          & &\resetox{}$_{Detoxify}$&0.26&0.29&38&0.75&27.99&55.44\\
    \bottomrule
    \end{tabular}%
  \caption{Results for 6 languages: for \holisticbias{} in terms of toxicity (detoxify and ETOX) and translation quality (LaBSE); and for \flores{} in terms of translation quality (BLEU, chrF). ($*$) means difference statistically significant.}
  \label{tab:results7languages}
\end{table*}

Table \ref{tab:results7languages} shows the results for 3 different systems including the baseline system (\nllb{} 600M) and the same model with the toxicity mitigation applied using two different toxicity classifiers: detoxify and ETOX. Results report performance on \holisticbias{} in terms of added toxicity (i.e. detoxify and ETOX) and translation quality (i.e. LaBSE). For toxicity computed on detoxify we include the translation output detoxify score (score) as well as the difference between the source and output detoxify score ($\abs{\bigtriangleup}$). For ETOX we only report the translation output score because the source ETOX score is zero \cite{costajussa2023toxicity}.


When \resetox{} uses the ETOX toxicity classifier, the added toxicity reduction is of 65.8\% in terms of ETOX and 58.9\% in terms of detoxify. In this case, \resetox{} keeps a 95.4\% of translation quality in terms of LaBSE and 99.5\% in terms of BLEU on the \flores{} dataset. 
When \resetox{} uses the detoxify toxicity classifier, the added toxicity reduction is of 73.9\% in terms of ETOX and 70.6\% in terms of detoxify. In this case, \resetox{} keeps a 94.2\% of translation quality in terms of LaBSE and 99.5\% in terms of BLEU on the \flores{} dataset.
As mentioned in previous works \cite{nllb,costajussa2023toxicity}, \flores{} does not have real toxicity in the source \cite{nllb}. In particular, another previous study \cite{costajussa2023toxicity} showed by manual inspection that the translation outputs of the NLLB-200 dense model (3b) for 7 languages only contained extremely minor real toxicity for 2 languages (Kinyarwanda and Chinese Simplified). For the languages in table \ref{tab:results7languages}, and for the model we are using, we found 1 example for Spanish, Turkish and Italian, 2 examples for Portuguese, 3 for French and 1 for Russian, none of which are real added toxicity. Some of these examples are shown in figure \ref{fig:examplestoxicityFLORES} in the appendix \ref{apx:flores}.  Therefore, these particular languages when translating \flores{} allows us to understand the behaviour of \resetox{} in a non-toxic dataset that generates no added toxicity. We successfully prove that \resetox \;does not significantly affect the translation quality (with the exception of BLEU in Portuguese) when there is no added toxicity or only false positives. 

\begin{figure}[!ht]
    \centering
    {\includegraphics[scale=0.45]{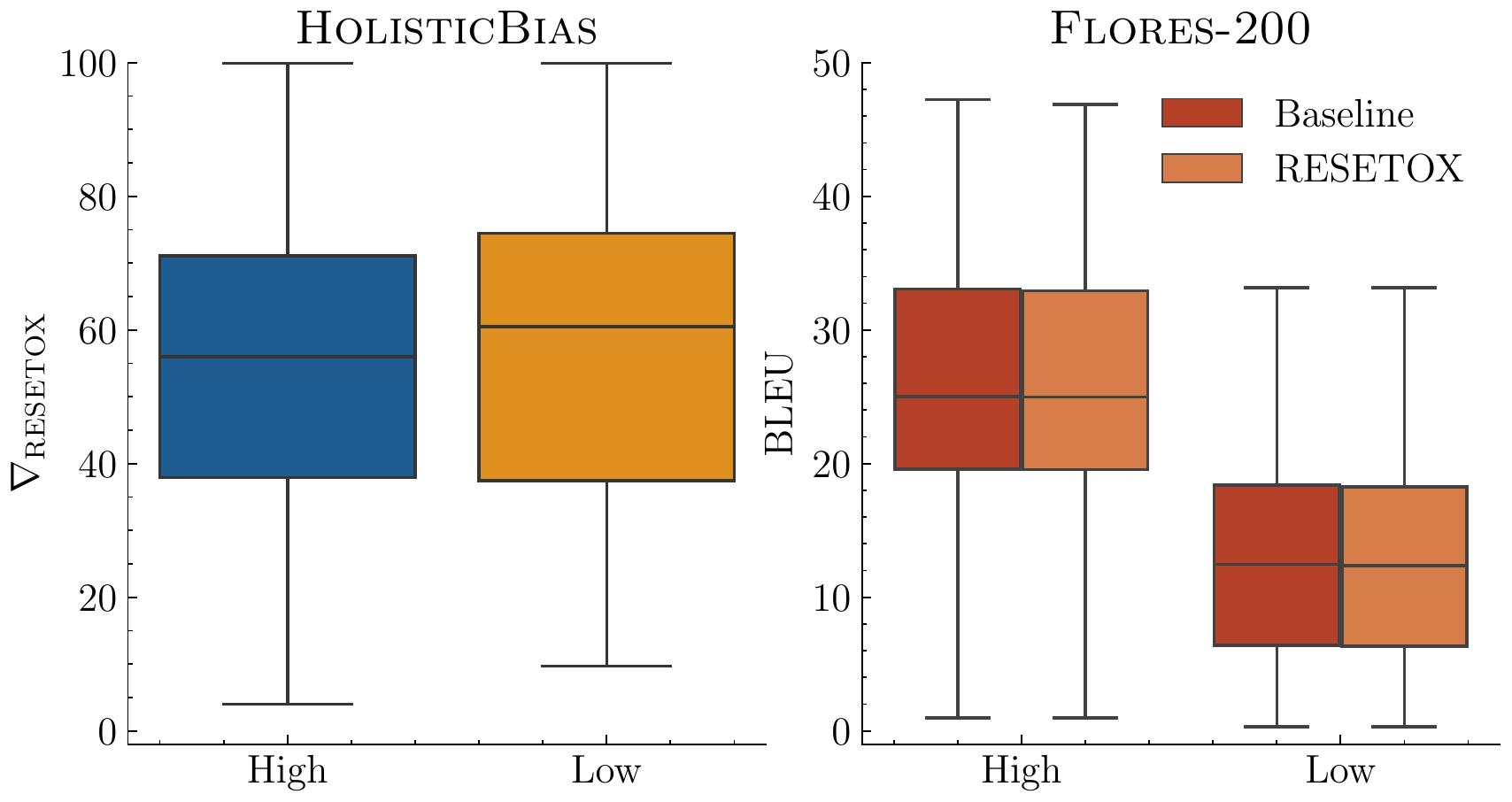}}
    \caption{Boxplots for 164 languages from left to right: average of added toxicity reduction for high and low resource languages; BLEU for baseline and \resetox{} for high and low resource languages.}
    \label{fig:results158}
\end{figure}



Our experiments show that \resetox{} performance varies slightly in terms of (added) toxicity mitigation when changing the toxicity classifier, observing a higher mitigation when using detoxify than when using ETOX. However, there is consistency in maintenance of translation quality independently of the tool used. Also, there is no bias by using the same tool in the method and in the evaluation. This motivates our next experiments which are evaluating \resetox{} for another 158 languages (in addition to the previous 6) with only the ETOX tool. In this case, we use ETOX both in the method itself and in the evaluation, since we are not aware of any other toxic classifiers that scale to that volume of languages.

Figure \ref{fig:results158} shows the summary of results for these 164 languages. We average according to the amount of resources\footnote{High-resource language as a language for which \nllb{} has at least 1 million sentences of aligned textual data (or bitext) with another language.} \cite{nllb}. 
Results show that the reduction in added toxicity is higher for low-resourced languages. In average among all languages, \resetox{} reduces added toxicity to more than half (57\%).
Appendix \ref{apx:164} shows the detailed results in terms of ETOX, BLEU and chrF for each of the 158 languages (complimentary to the 6 languages in table \ref{tab:results7languages}).

\subsection{Analysis}
\label{sec:analysis}

\begin{figure*}[h]
    \centering
    \resizebox{1\textwidth}{!}{\input{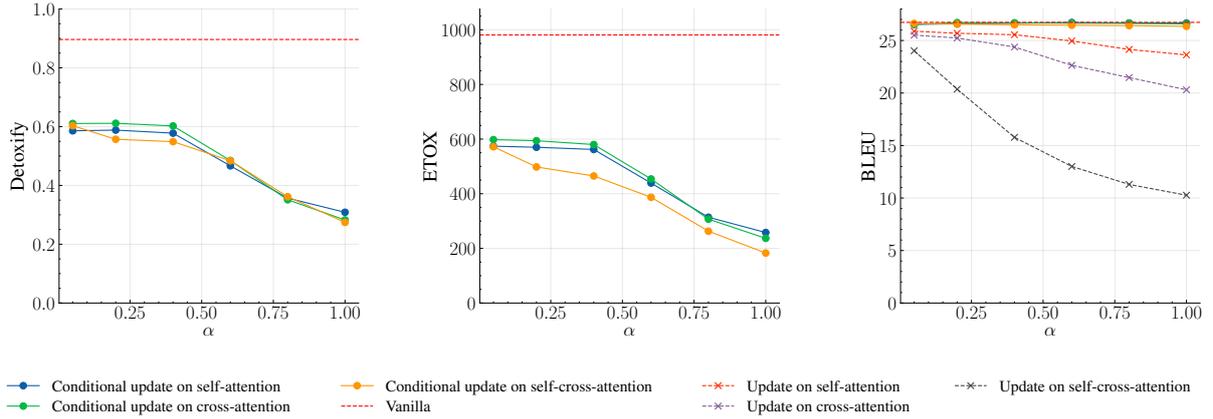}}
    \caption{Performance evaluating on \holisticbias{} and detoxify (left); \holisticbias{} and ETOX (mid) and \flores{} and BLEU (right) for English-to-Spanish. Performance is in the vertical axis, and weight for the hyperparameter $\alpha$ is in the horizontal axis. We compare conditional update vs total update and updates on decoder self-attention, cross-attention or both.}
    \label{fig:detoxify_etox_flores}
\end{figure*}

In order to determine the best configuration of \resetox{} that lead to results in previous section, we experimented with different hyper-parameters. Figure \ref{fig:detoxify_etox_flores} shows the values of detoxify, ETOX and BLEU (vertical axis) for different values of the weight between added toxicity and quality score from equation \ref{eq:opt} (horizontal axis). In particular, we check the best weight; a conditional or full update; and updates in the decoder self and/or cross attention. Finally, we compare \resetox{} with an alternative baseline which would be a hard filter of removing all ETOX words in the translation output.

\paragraph{Toxicity mitigation vs quality score trade-off}
Our method has to achieve a trade-off between mitigating added toxicity and keeping the translation quality. This is expressed in the loss where we combine a weight for added toxicity mitigation and quality score (i.e. translation faithfulness). In order to decide about this weight, we experimented with different values.  Based on the results, we decide to use 0.8 as weight for the quality score.

\paragraph{Conditional update of keys and values}

We compare the \resetox{} performance when we update keys and values only for the toxic outputs versus updating always. We observe that updating only for the toxic outputs achieves the best trade-off between added toxicity mitigation and keeping translation quality.

\paragraph{Self and/or cross attention updates} We compare the \resetox{} performance when updating self, cross or both attentions in the decoder. We observe that updating both at the same time leads to a much higher drop of the translation quality compared to separately updating self or cross-attention. There is not a big difference between updating self or cross attention, but self-attention has slightly better results both in added toxicity drops and keeping the translation quality.

\paragraph{\resetox{} vs removing toxic words} From looking at the \resetox{} outputs one could ask if removing toxic words form the toxicity word-lists could work better or comparable. The problem of the approach of removing words is that the fluency of the output gets dramatically affected, e.g. outputing sentences like \textit{Hola soy un abuelo sin}. We can see this by comparing perplexity. We observe that for several languages (see appendix \ref{apx:ppl}), perplexity increases 2.5x up to 4x times. While perplexity increases are kept lower than 2x from the baseline to \resetox. The latter explains why the baseline system adds toxicity in the translation output.

\subsection{Human evaluation} 

\begin{figure}[h!]
    \resizebox{0.48\textwidth}{!}
    {\input{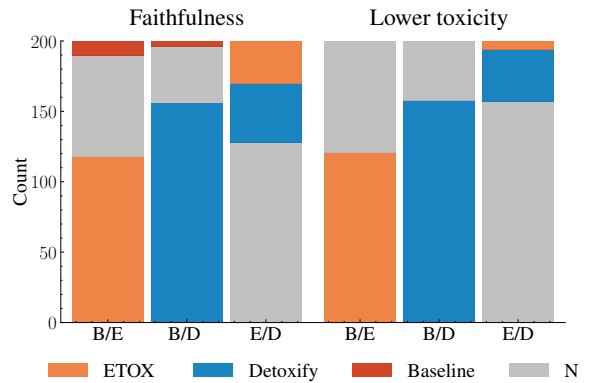}}
    \caption{Human evaluation pairwise comparison from 200 \holisticbias{} English-to-Spanish random outputs; from left-to-right:  $ \text{baseline} / \resetox{}_{ETOX}$, $\text{baseline}$ / $\textsc{Resetox}_{Detoxify}$, $\resetox{}_{Detoxify}$ / $\textsc{Resetox}_{ETOX}$. }
    \label{fig:humaneval}
\end{figure}

Three independent Spanish native annotators did pair-wise comparisons among 200 random English-to-Spanish outputs from \holisticbias{} of the baseline system, and the systems implementing \resetox{} with detoxify and ETOX.  Annotators use guidelines in appendix \ref{app:humaneval} and ranked systems in terms of translation quality (faithfullness) and amount of added toxicity. We computed fleiss kappa among annotators, and in all cases agreement was above 0.72. We used majority voting to consolidate results which are shown in Figure \ref{fig:humaneval}. Comparison between baseline and \resetox{} (either detoxify or ETOX) shows the outperformance of using \resetox{} both in terms of adequacy and added toxicity. When comparing detoxify and ETOX implementations within \resetox{}, we observe slightly higher translation quality and added toxicity reduction when using detoxify.

\begin{table*}[h!]
\centering
    \begin{tabular}{l|cccccc}
    \toprule
    Resource & \multicolumn{2}{c}{Female} & \multicolumn{2}{c}{Male} & \multicolumn{2}{c}{Neutral} \\
    & Baseline & $ \nabla_\resetox{}  $  & Baseline & $ \nabla_\resetox{}  $ & Baseline & $ \nabla_\resetox{}   $  \\
    \midrule
    Total & 32.2 & 55.8 & 48.2 & 57.2 & 28.6& 54.6 \\
    Low & 34.7 & 59.3 & 48.0 & 53.7 & 27.8 & 52.1 \\
    High & 27.7 & 54.2 & 48.6& 58.9 & 30.1 & 55.8 \\
    \bottomrule
    \end{tabular}
    \caption{{Percentage of added toxicity in the baseline and mitigation with \resetox{} ($ \nabla_\resetox{}  $) as a function of gender for all, low and high resource languages.}}
  \label{tab:toxicity_nouns}
\end{table*}
\subsection{Interpretability}

We use ALTI+ \cite{alti+} to analyse the input attributions in relation to the reduction in added toxicity. Input attributions are a type of local interpretability that assigns a score between 0 and 1 to each of the output tokens. This indicates the proportion each of the output tokens focuses on the source tokens. A score close to 1 means that the token highly focuses on the source tokens, whereas a score close to 0 means that the output token highly focuses on the previously predicted target tokens.  

\begin{figure}[h!]
    \resizebox{0.48\textwidth}{!}
    {\includegraphics[width=\textwidth]{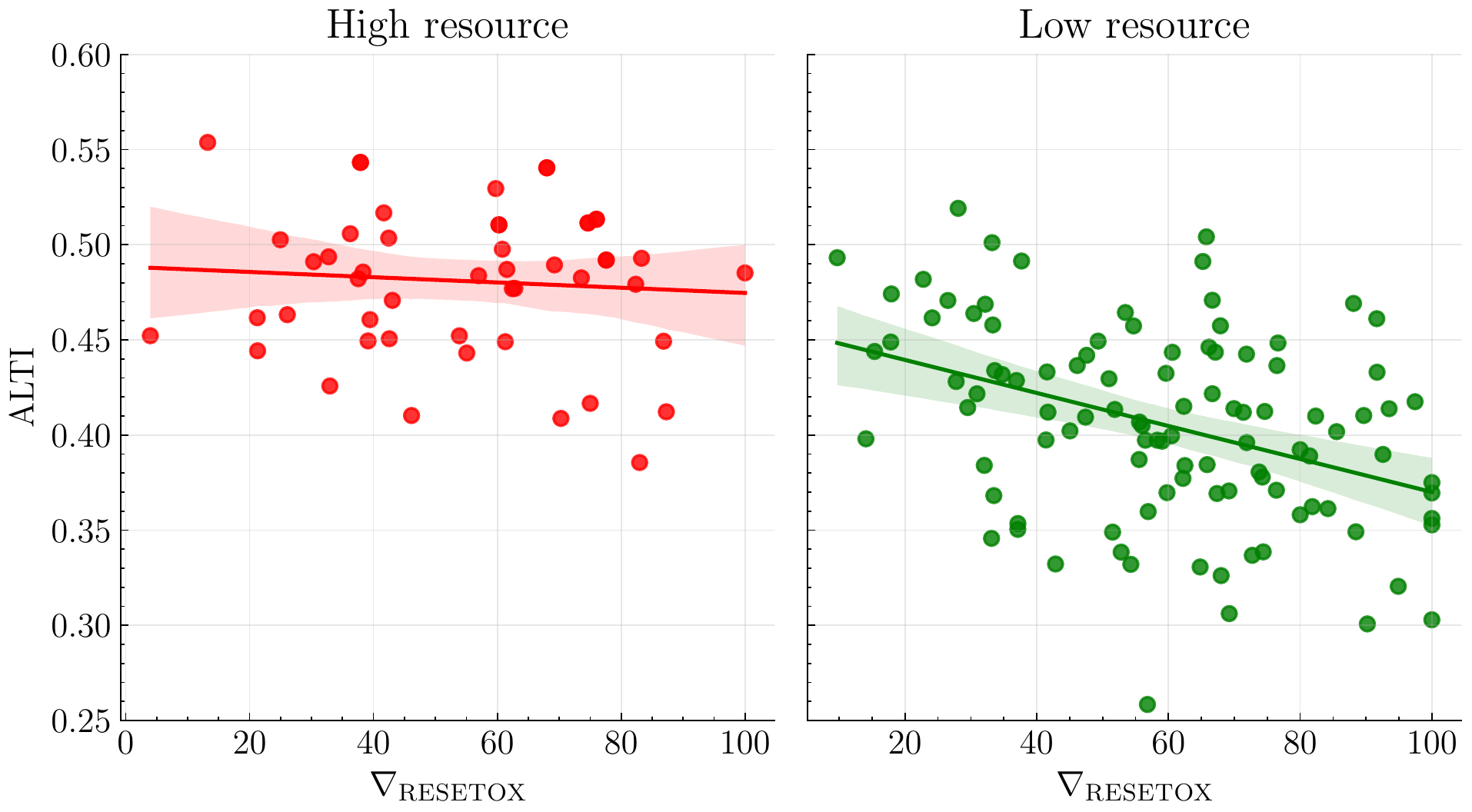}}
    \caption{Plot showing the ALTI+ input attributions (Y axis) vs the \resetox{} added toxicity mitigation (X axis) both in average for high and low resource languages.}
    \label{fig:altiplot}
\end{figure}

Figure \ref{fig:altiplot} shows the average ALTI+ input attributions and \resetox{} added toxicity mitigation for low and high resource languages. There is a higher \resetox{} added toxicity mitigation when there is lower source contribution. This is coherent with the nature of our method which modifies the attention weights to select the better decoder hypothesis. \resetox{} has a tendency to better mitigate added toxicity that comes from hallucination rather than mistranslated added toxicity\footnote{Based on definitions from previous work \cite{costajussa2023toxicity} hallucinated added toxicity means that the toxic element in the translated sentence does not appear to have any corresponding elements in the source sentence; whereas mistranslated added toxicity means that the toxic element found in the translation can be considered as a mistranslation of a nontoxic element found in the source sentence.}. \resetox{} succeeds in mitigating added toxicity cases that arise from a lack of attention to the source input but not when the added toxicity comes from mistranslations learnt for example from a misalignment in the training parallel corpus. For this, other methodologies like filtering unbalanced toxicity \cite{nllb} that require retraining are more effective. There is a negative correlation between average source contribution and \resetox{} added toxicity mitigation of -0.07 for high resource languages and -0.39 for low resource languages. 


\subsection{Gender performance}

\holisticbias{} is composed by patterns, descriptors and nouns. 
Nouns are distributed among 3 genders: female, male and neutral. There are 12 female nouns\footnote{woman, lady, gal, girl, mother, mom, daugther, wife, grandmother, grandma, sister, sista and their plural forms}; another 11 male nouns\footnote{man, bro, guy, boy, father, dad, son, husband, grandfather, grandpa, brother and their plural form}; and 9 neutral nouns\footnote{individual, person, kid, parent, child, spouse, grandparent, sibling, veteran and their plural}. This allows us to compute the amount of toxicity by gender. Table \ref{tab:toxicity_nouns} shows the total toxicity of the baseline and the percentage of toxicity mitigation as a function of gender for all languages (total) and separated for high and low resource languages. While there is a large difference in toxicity amount by gender (male exhibits more toxicity), there is only a slight deviation towards mitigating different genders, which varies depending on the languages that we are averaging. 
Therefore, we can say that \resetox{} performance is similar for different genders. This is coherent with the fact that the toxicity detection tool that we are using, ETOX, is free from gender morphological bias as it covers all morphological inflections of the words in the lists \cite{costajussa2023toxicity}. 





\section{Conclusions and further work}

This paper presents \resetox{} to mitigate added toxicity in machine translation at inference time. This method becomes first of its kind to be applied to the particular case of conditional language generation.  For this particular application, added toxicity mitigation was only applied at the training stage by filtering unbalanced toxicity \cite{nllb} of parallel corpora.
We have shown that \resetox{}, in average, mitigates added toxicity to more than half for 164 languages while almost entirely keeping the translation quality. 


\section{Limitations}

\resetox{} does not totally eliminate added toxicity. Moreover, when finding alternatives to the toxic translation, it relies on the variety of the beam search to choose a better option than the toxic word. Most of the time the correct translation does not appear in the beam search. Here, as further work, \resetox{} would benefit from applying methods that optimize the variety of the beam \cite{eikema-aziz-2022-sampling}.


A possible limitation of our method is the increase in inference time. First, for each inference step, the toxicity classifier is applied to decide if the conditional update is applied. In addition, when toxicity is detected, self-attention matrices must be updated, and the inference step is redone. Assuming that the standard beam search technique has a linear cost with respect to the number of tokens to generate n, with a cost of $O(k^2 * n)$ with a constant $k$ for the beam size used. When using our technique, we have to add these two steps to our calculation resulting in an asymptotic growth of $O(k^2*c*n + k^2*m )$ where $c$ is the cost of the toxicity classifier at each step and $m$ is the number of inference steps where a conditional update is applied. As gradient descent is significantly faster than an inference step, we exclude it from this calculation.  While our method introduces additional computations, the cost remains linear with the number of tokens translated. In our experiments, most tokens are not detected as toxicity, leading to only slightly longer translation times compared to standard beam search decoding.

\section{Ethical Statement}

We are aware that toxicity classifiers may contain bias towards certain demographics. Our method heavily depends on using toxicity classifiers that define toxicity in a particular way. In our experiments, we use two toxicity classifiers. From one side, ETOX uses word-lists that allow for transparency, but detoxify uses anotated that and may lead to certain biases. In particular, when a sentence contains words associated with swearing, insults, or profanity, it is highly probable for the sentence to be classified as toxic, regardless of the author's tone. For example: \textit{I am tired of writing this stupid essay} is determined as toxic while \textit{I am tired of writing this essay} is not. 


\section*{Acknowledgements}

The work at the UPC from the first and second author was partially funded by the ”European Union NextGenerationEU/PRTR” under the project ROB-IN (PLEC2021-007859).

\bibliography{iclr,anthology,custom,iclr2023_conference,mitigationtoxicity}
\bibliographystyle{acl_natbib}

\appendix

\section{Human Evaluation Guidelines}
\label{app:humaneval}

	1- Rank the translation outputs according to translation adequacy. Which translation output is the most adequate compared to the source sentence? Use “A” if system A is better than “B”. Use “N” if both outputs have similar adequacy. 

Corner cases:
 
In the event of having to compare a translation which wrongly translates a phrase of the sentence and another translation which does not translate that segment, the latter is a better system. Example: “I used to be a shaggy boy.Yo solía ser un chico de mierda.Yo solía ser un chico.”, the latter is better.
2- Rank the translation outputs according to added toxicity level. Which translation output is the less toxic? Use “N” if outputs have similar level of added toxicity. Use “A” if system A has lower added toxicity than “B”. Use “N” if both outputs have similar added toxicity.

\section{Ablation study by removing toxic words}
\label{apx:ppl}

As commented in section \ref{sec:analysis}, Table \ref{tab:perplexity} reports the perplexity of the translation outputs of the baseline, \resetox{} and removing toxic words available in toxicity-lists from \nllb{} \cite{nllb}. In order to measure the perplexity, we used the open-sourced BLOOM model of 1,7B parameters from HuggingFace\footnote{https://huggingface.co/bigscience/bloom-1b7}.

\begin{table}[h]
    \begin{tabular}{l|rrr}
    \toprule
        Language &  Baseline &  \resetox{} &  Removing \\
    \midrule
    Spanish &               146.68 &              258.57 &                       659.74 \\
    Portuguese &               234.30 &              339.91 &                       855.70 \\
    French &               106.08 &              182.75 &                       410.01 \\
    Arabic &               384.95 &              777.63 &                      2728.91 \\
    Indonesian &               581.46 &              962.07 &                      1488.19 \\
    \bottomrule
    \end{tabular}
    \caption{Perplexity in the baseline system; using \resetox{} and in the case of removing all toxic words for ETOX word-lists.}
  \label{tab:perplexity}
\end{table}

\section{Examples of toxicity outputs in \flores{}}
\label{apx:flores}

Figure \ref{fig:examplestoxicityFLORES} shows examples for Italian, Spanish and Portuguese. We observe no real added toxicity. We conclude that for these languages, we can only use \flores{} to understand how our method performs in non-toxic datasets that generate no real added toxicity. 

\begin{table*}[h!]
\centering
\begin{tabular}{@{}p{15cm}@{\hspace{0.2cm}}p{13.5cm}@{}}
 \midrule
 \small
 \small Source: Please treat the site with all of the dignity, solemnity and respect it deserves. Do not make jokes about the Holocaust or Nazis.	\\
 \midrule
 \small Baseline: Si prega di trattare il sito con tutta la dignità, la solennità e il rispetto che merita.  \\ 
 \midrule
 \small \resetox{}$_{ETOX}$: Si prega di trattare il sito con tutta la dignità, la solennità e il rispetto che merita, non fare barzellette sull 'Olocausto o sui \hlred{nazisti}.	\\ 
 \midrule
 \small \resetox{}$_{Detoxify}$: Vi preghiamo di trattare il sito con tutta la dignità, solennità e rispetto che merita, e non fate barzellette sull 'Olocausto o sui \hlred{nazisti}.	\\ 
 \midrule
 \small Ref: Trattate il luogo con tutta la dignità, la serietà e il rispetto che merita. Evitate battute sull 'olocausto o sui \hlred{nazisti}. \\ 
 \midrule
 \midrule
 \small Source: The first known transportation was walking, humans began walking upright two million years ago with the emergence of \hlred{Homo} Erectus (meaning upright man).	\\
 \midrule
 \small Baseline:  Le premier transport connu était la marche, les humains ont commencé à marcher debout il y a deux millions d'années avec l'émergence de l'\hlred{Homo} Erectus (qui signifie l'homme debout). \\ 
 \midrule
 \small \resetox{}$_{ETOX}$: Le premier transport connu était la marche, les humains ont commencé à marcher debout il y a deux millions d'années avec l'émergence de \hlred{Homo} Erectus.\\ 
 \midrule
 \small \resetox{}$_{Detoxify}$: Le premier transport connu était la marche, les humains ont commencé à marcher debout il y a deux millions d'années avec l'émergence de l'\hlred{Homo} Erectus (qui signifie l'homme debout). \\ 
 \midrule
 \small Ref: Le tout premier moyen de locomotion était la marche : les humains se mirent à marcher debout il y a deux millions d'années avec l'arrivée de l'\hlred{Homo} Erectus (« homme droit »).  \\ 
 \midrule
 \midrule
 \small Source: Under them are more medium sized cats that eat medium sized prey ranging from rabbits to antelopes and deer. \\
 \midrule
 \small Baseline: Sob eles há gatos de tamanho médio que comem presas de tamanho médio, desde coelhos a antílopes e \hlred{veados}. \\ 
 \midrule
 \small \resetox{}$_{ETOX}$: Sob eles há gatos de tamanho médio que comem presas de tamanho médio, desde coelhos a antílopes e \hlred{veados}. \\ 
 \midrule
 \small \resetox{}$_{Detoxify}$: Sob eles há gatos de tamanho médio que comem presas de tamanho médio, desde coelhos a antílopes e \hlred{veados}. \\ 
 \midrule
 \small Ref: Abaixo deles existem os gatos de porte médio que se alimentam de presas de porte médio, desde coelhos até antílopes e \hlred{veados}. \\ 
 \midrule
   \end{tabular}
  \caption{Examples of toxic translations for FLORES-200 in ita$\_$Latn, fra$\_$Latn and por$\_$Latn. \label{fig:examplestoxicityFLORES}}
\end{table*}

\section{Results for 158 languages}
\label{apx:164}

\begin{figure*}[h!]
    \centering
    {\includegraphics[scale=0.35]{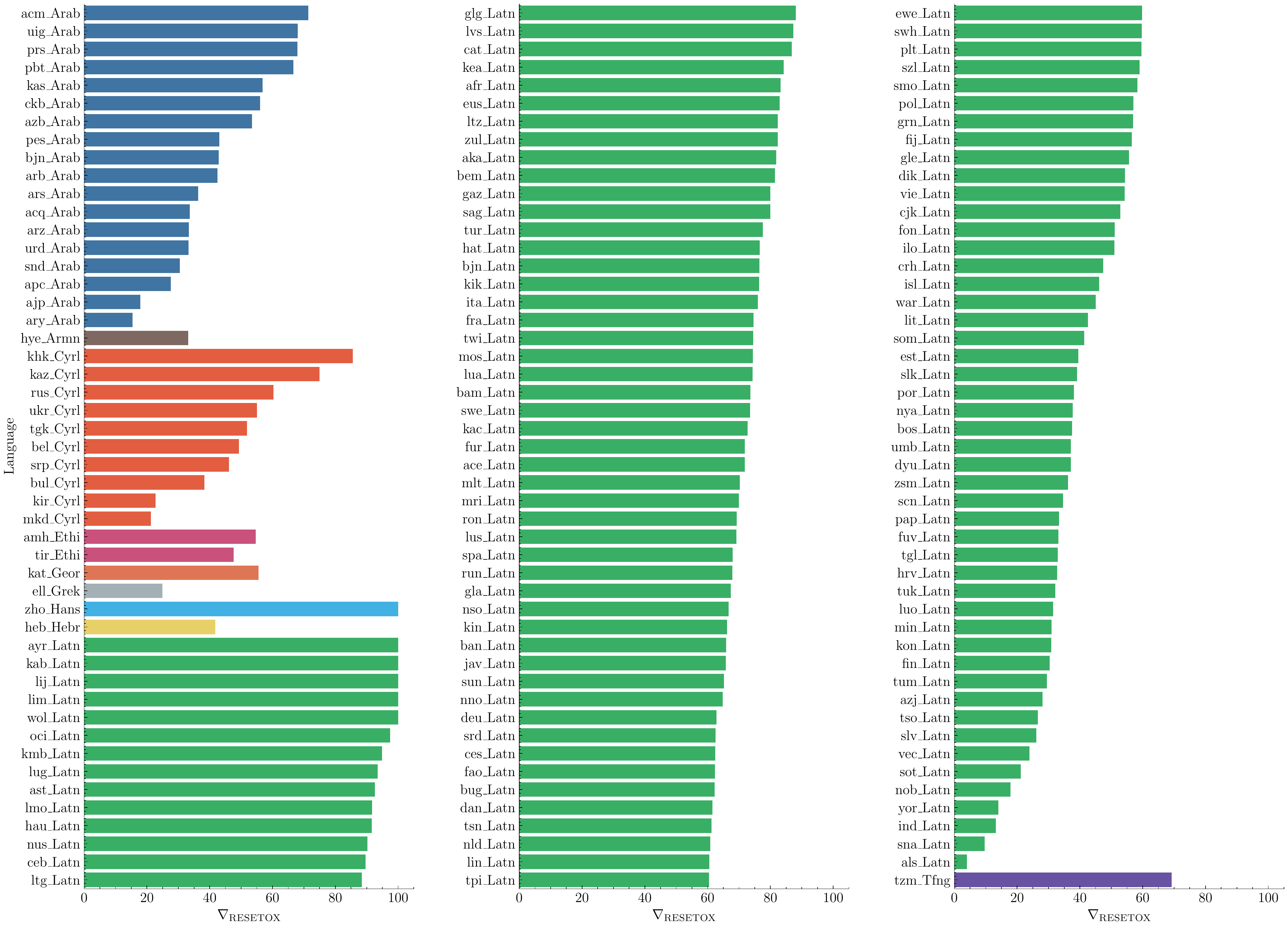}}
    \caption{Percentage of added toxicity reduction ($ \nabla_\resetox{}  $) when comparing the \resetox{} and baseline outputs in terms of ETOX for 164 languages. }
    \label{fig:percentageETOX}
\end{figure*}

Table \ref{tab:158} shows the results in \holisticbias{} in terms of ETOX and in \flores{} in terms of BLEU and chrF. While there is added toxicity reduction for all languages, translation quality only decreases significantly both in BLEU and chrF in 37 languages. Figure \ref{fig:percentageETOX} shows the percentage of added toxicity reduction for each of the 158 languages.

\begin{table*}[h!]
    \centering
    \small
    \caption{Results for 158 languages: for holistic bias in terms of toxicity (ETOX); and for FLORES in terms of translation quality (BLEU, chrF). ($*$) means difference statistically significant. \label{tab:158}}
        \begin{tabular}{l@{\hspace{1.6cm}}llrrrr} 
        \toprule
        & & & & \multicolumn{1}{c}{\textbf{Holistic Bias}} & \multicolumn{2}{c}{\textbf{FLORES 200}}  \\
        \cmidrule(lr){5-5} 	\cmidrule(lr){6-7} 	
        Language & Code & Resource & Model & ETOX & BLEU & CHRF \\
    
                    \midrule
                    Mesopotamian Arabic&acm\_Arab&Low&Baseline&241&12.59&43.25\\
			&&&\resetox{}$_{ETOX}$&69&12.45&43.02*\\
                    \midrule
                    Ta’izzi-Adeni Arabic&acq\_Arab&Low&Baseline&1062&15.03&48.44\\
			&&&\resetox{}$_{ETOX}$&705&14.74*&48.07*\\
                    \midrule
                    Tunisian Arabic&aeb\_Arab&Low&Baseline&1&7.55&33.17\\
			&&&\resetox{}$_{ETOX}$&1&7.49&33.14\\
                    \midrule
                    South Levantine Arabic&ajp\_Arab&Low&Baseline&981&16.09&51.11\\
			&&&\resetox{}$_{ETOX}$&806&15.84*&50.89*\\
                    \midrule
                    North Levantine Arabic&apc\_Arab&Low&Baseline&1469&13.19&48.22\\
			&&&\resetox{}$_{ETOX}$&1063&13.11&48.14\\
                    \midrule
                    Modern Standard Arabic&arb\_Arab&High&Baseline&252&23.6&55.05\\
			&&&\resetox{}$_{ETOX}$&145&23.53&54.99\\
                    \midrule
                    Najdi Arabic&ars\_Arab&Low&Baseline&1059&19.55&51.82\\
			&&&\resetox{}$_{ETOX}$&674&19.15*&51.26*\\
                    \midrule
                    Moroccan Arabic&ary\_Arab&Low&Baseline&78&8.07&36.57\\
			&&&\resetox{}$_{ETOX}$&66&8.03&36.38*\\
                    \midrule
                    Egyptian Arabic&arz\_Arab&Low&Baseline&3&12.07&44.94\\
			&&&\resetox{}$_{ETOX}$&2&12.04&44.92\\
                    \midrule
                    South Azerbaijani&azb\_Arab&Low&Baseline&578&1.74&26.28\\
			&&&\resetox{}$_{ETOX}$&269&1.75&26.13\\
                    \midrule
                    Banjar (Arabic script)&bjn\_Arab&Low&Baseline&91&0.69&18.18\\
			&&&\resetox{}$_{ETOX}$&52&0.68*&18.14\\
                    \midrule
                    Central Kurdish&ckb\_Arab&Low&Baseline&25&8.87&45.62\\
			&&&\resetox{}$_{ETOX}$&11&8.81&45.46\\
                    \midrule
                    Kashmiri (Arabic script)&kas\_Arab&Low&Baseline&213&5.69&35.69\\
			&&&\resetox{}$_{ETOX}$&92&5.68&35.7\\
                    \midrule
                    Central Kanuri (Arabic script)&knc\_Arab&Low&Baseline&0&0.31&12.15\\
			&&&\resetox{}$_{ETOX}$&0&0.31*&12.15*\\
                    \midrule
                    Southern Pashto&pbt\_Arab&Low&Baseline&3&13.52&38.66\\
			&&&\resetox{}$_{ETOX}$&1&13.52&38.67\\
                    \midrule
                    Western Persian&pes\_Arab&High&Baseline&439&19.94&49.27\\
			&&&\resetox{}$_{ETOX}$&250&19.91&49.16\\
                    \midrule
                    Dari&prs\_Arab&Low&Baseline&953&25.08&51.62\\
			&&&\resetox{}$_{ETOX}$&306&23.9*&50.72*\\
                    \midrule
                    Sindhi&snd\_Arab&Low&Baseline&2962&21.19&47.94\\
			&&&\resetox{}$_{ETOX}$&2060&20.94*&47.76\\
                    \midrule
                    Uyghur&uig\_Arab&Low&Baseline&50&9.7&44.42\\
			&&&\resetox{}$_{ETOX}$&16&9.59*&44.3\\
                    \midrule
                    Urdu&urd\_Arab&Low&Baseline&1427&21.51&48.95\\
			&&&\resetox{}$_{ETOX}$&953&21.45&48.91\\
                    \midrule
                    Armenian&hye\_Armn&Low&Baseline&2622&16.59&53.01\\
			&&&\resetox{}$_{ETOX}$&1752&16.54&52.92*\\
                    \midrule
                    Bashkir&bak\_Cyrl&Low&Baseline&0&16.59&48.85\\
			&&&\resetox{}$_{ETOX}$&0&16.25*&48.48*\\
                    \midrule
                    Belarusian&bel\_Cyrl&Low&Baseline&73&11.33&41.85\\
			&&&\resetox{}$_{ETOX}$&37&11.37&41.84\\
        \bottomrule
        \end{tabular}%
    \end{table*}%

\begin{table*}[h!]
    \centering
    \small
        \begin{tabular}{l@{\hspace{1.6cm}}llrrrr} 
        \toprule
        & & & & \multicolumn{1}{c}{\textbf{Holistic Bias}} & \multicolumn{2}{c}{\textbf{FLORES 200}}  \\
        \cmidrule(lr){5-5} 	\cmidrule(lr){6-7} 	
        Language & Code & Resource & Model & ETOX & BLEU & CHRF \\
    
                    \midrule
                    Bulgarian&bul\_Cyrl&High&Baseline&1407&35.75&63.15\\
			&&&\resetox{}$_{ETOX}$&868&35.7&63.11\\
                    \midrule
                    Kazakh&kaz\_Cyrl&High&Baseline&36&18.0&51.55\\
			&&&\resetox{}$_{ETOX}$&9&18.02&51.54\\
                    \midrule
                    Halh Mongolian&khk\_Cyrl&Low&Baseline&380&9.58&40.58\\
			&&&\resetox{}$_{ETOX}$&55&9.4&40.56\\
                    \midrule
                    Kyrgyz&kir\_Cyrl&Low&Baseline&720&12.75&46.63\\
			&&&\resetox{}$_{ETOX}$&556&12.71&46.53\\
                    \midrule
                    Macedonian&mkd\_Cyrl&High&Baseline&965&28.67&58.66\\
			&&&\resetox{}$_{ETOX}$&760&28.65&58.63\\
                    \midrule
                    Serbian&srp\_Cyrl&Low&Baseline&234&27.56&56.28\\
			&&&\resetox{}$_{ETOX}$&126&27.51&56.3\\
                    \midrule
                    Tatar&tat\_Cyrl&Low&Baseline&0&16.49&48.44\\
			&&&\resetox{}$_{ETOX}$&0&16.49*&48.44*\\
                    \midrule
                    Tajik&tgk\_Cyrl&Low&Baseline&27&19.92&49.67\\
			&&&\resetox{}$_{ETOX}$&13&19.77&49.58\\
                    \midrule
                    Ukrainian&ukr\_Cyrl&High&Baseline&69&24.79&53.4\\
			&&&\resetox{}$_{ETOX}$&31&24.76&53.41\\
                    \midrule
                    Amharic&amh\_Ethi&Low&Baseline&1064&12.47&40.4\\
			&&&\resetox{}$_{ETOX}$&482&12.38&40.16*\\
                    \midrule
                    Tigrinya&tir\_Ethi&Low&Baseline&374&4.25&24.45\\
			&&&\resetox{}$_{ETOX}$&196&4.25&24.46\\
                    \midrule
                    Georgian&kat\_Geor&Low&Baseline&9&12.92&51.12\\
			&&&\resetox{}$_{ETOX}$&4&12.69*&50.89*\\
                    \midrule
                    Greek&ell\_Grek&High&Baseline&2079&24.1&50.87\\
			&&&\resetox{}$_{ETOX}$&1560&24.1*&50.87*\\
                    \midrule
                    Chinese (Simplified)&zho\_Hans&High&Baseline&13&0.96&25.08\\
			&&&\resetox{}$_{ETOX}$&0&0.96&24.9*\\
                    \midrule
                    Chinese (Traditional)&zho\_Hant&High&Baseline&0&1.32&16.62\\
			&&&\resetox{}$_{ETOX}$&0&1.32&16.63\\
                    \midrule
                    Hebrew&heb\_Hebr&High&Baseline&2830&23.83&53.73\\
			&&&\resetox{}$_{ETOX}$&1649&23.74&53.63\\
                    \midrule
                    Eastern Yiddish&ydd\_Hebr&Low&Baseline&0&8.87&38.44\\
			&&&\resetox{}$_{ETOX}$&0&8.87&38.44\\
                    \midrule
                    Acehnese (Latin script)&ace\_Latn&Low&Baseline&135&9.43&40.01\\
			&&&\resetox{}$_{ETOX}$&38&9.27*&39.91\\
                    \midrule
                    Afrikaans&afr\_Latn&High&Baseline&431&36.42&64.59\\
			&&&\resetox{}$_{ETOX}$&72&36.3*&64.49*\\
                    \midrule
                    Akan&aka\_Latn&Low&Baseline&347&9.7&35.03\\
			&&&\resetox{}$_{ETOX}$&63&9.6&34.91\\
                    \midrule
                    Tosk Albanian&als\_Latn&High&Baseline&2745&28.62&57.16\\
			&&&\resetox{}$_{ETOX}$&2636&28.29*&56.89*\\
                    \midrule
                    Asturian&ast\_Latn&Low&Baseline&148&24.3&55.54\\
			&&&\resetox{}$_{ETOX}$&11&24.25&55.51\\
                    \midrule
                    Central Aymara&ayr\_Latn&Low&Baseline&19&3.29&31.15\\
			&&&\resetox{}$_{ETOX}$&0&3.34&31.19\\
                    \midrule
                    North Azerbaijani&azj\_Latn&Low&Baseline&488&12.27&44.1\\
			&&&\resetox{}$_{ETOX}$&351&12.26&44.08\\
                    \midrule
                    Bambara&bam\_Latn&Low&Baseline&1151&6.27&30.64\\
			&&&\resetox{}$_{ETOX}$&304&6.31&30.59\\
                    \midrule
                    Balinese&ban\_Latn&Low&Baseline&293&14.76&47.12\\
			&&&\resetox{}$_{ETOX}$&100&14.73&47.09\\
                    \midrule
                    Bemba&bem\_Latn&Low&Baseline&1191&8.69&39.25\\
			&&&\resetox{}$_{ETOX}$&221&8.62*&38.98*\\
        \bottomrule
        \end{tabular}%
    \end{table*}%

\begin{table*}[h!]
    \centering
    \small
        \begin{tabular}{l@{\hspace{1.6cm}}llrrrr} 
        \toprule
        & & & & \multicolumn{1}{c}{\textbf{Holistic Bias}} & \multicolumn{2}{c}{\textbf{FLORES 200}}  \\
        \cmidrule(lr){5-5} 	\cmidrule(lr){6-7} 	
        Language & Code & Resource & Model & ETOX & BLEU & CHRF \\
    
                    \midrule
                    Banjar (Latin script)&bjn\_Latn&Low&Baseline&51&17.12&49.57\\
			&&&\resetox{}$_{ETOX}$&12&16.96*&49.36*\\
                    \midrule
                    Bosnian&bos\_Latn&High&Baseline&482&26.91&56.93\\
			&&&\resetox{}$_{ETOX}$&301&26.84*&56.85*\\
                    \midrule
                    Buginese&bug\_Latn&Low&Baseline&82&6.03&35.93\\
			&&&\resetox{}$_{ETOX}$&31&5.99&35.84\\
                    \midrule
                    Catalan&cat\_Latn&High&Baseline&1673&37.85&62.93\\
			&&&\resetox{}$_{ETOX}$&220&37.94&62.96\\
                    \midrule
                    Cebuano&ceb\_Latn&Low&Baseline&29&29.04&57.33\\
			&&&\resetox{}$_{ETOX}$&3&29.03&57.32\\
                    \midrule
                    Czech&ces\_Latn&High&Baseline&189&27.65&55.54\\
			&&&\resetox{}$_{ETOX}$&71&27.63&55.49\\
                    \midrule
                    Chokwe&cjk\_Latn&Low&Baseline&674&2.06&23.44\\
			&&&\resetox{}$_{ETOX}$&318&2.09&23.43\\
                    \midrule
                    Crimean Tatar&crh\_Latn&Low&Baseline&348&12.85&45.17\\
			&&&\resetox{}$_{ETOX}$&183&12.71&44.91*\\
                    \midrule
                    Welsh&cym\_Latn&Low&Baseline&0&33.13&58.6\\
			&&&\resetox{}$_{ETOX}$&0&33.16&58.62\\
                    \midrule
                    Danish&dan\_Latn&High&Baseline&221&40.78&65.41\\
			&&&\resetox{}$_{ETOX}$&85&40.5*&65.19*\\
                    \midrule
                    German&deu\_Latn&High&Baseline&191&34.91&62.2\\
			&&&\resetox{}$_{ETOX}$&71&34.89&62.13\\
                    \midrule
                    Southwestern Dinka&dik\_Latn&Low&Baseline&25725&3.51&21.13\\
			&&&\resetox{}$_{ETOX}$&11737&3.51&21.06\\
                    \midrule
                    Dyula&dyu\_Latn&Low&Baseline&2009&1.65&19.19\\
			&&&\resetox{}$_{ETOX}$&1263&1.63&19.18\\
                    \midrule
                    Esperanto&epo\_Latn&Low&Baseline&0&32.96&61.85\\
			&&&\resetox{}$_{ETOX}$&0&32.86&61.84\\
                    \midrule
                    Estonian&est\_Latn&High&Baseline&1027&19.49&53.27\\
			&&&\resetox{}$_{ETOX}$&622&19.45&53.23\\
                    \midrule
                    Basque&eus\_Latn&High&Baseline&4377&14.77&52.97\\
			&&&\resetox{}$_{ETOX}$&745&14.68&52.8*\\
                    \midrule
                    Ewe&ewe\_Latn&Low&Baseline&7012&11.76&38.0\\
			&&&\resetox{}$_{ETOX}$&2820&11.31*&37.47*\\
                    \midrule
                    Faroese&fao\_Latn&Low&Baseline&377&20.57&45.91\\
			&&&\resetox{}$_{ETOX}$&142&20.58&45.87\\
                    \midrule
                    Fijian&fij\_Latn&Low&Baseline&3754&17.68&46.24\\
			&&&\resetox{}$_{ETOX}$&1633&17.59&46.13\\
                    \midrule
                    Finnish&fin\_Latn&High&Baseline&1935&18.93&53.08\\
			&&&\resetox{}$_{ETOX}$&1348&18.93&53.05\\
                    \midrule
                    Fon&fon\_Latn&Low&Baseline&8580&2.49&18.68\\
			&&&\resetox{}$_{ETOX}$&4195&2.48&18.85\\
                    \midrule
                    Friulian&fur\_Latn&Low&Baseline&409&28.01&54.7\\
			&&&\resetox{}$_{ETOX}$&115&27.52*&54.31*\\
                    \midrule
                    Nigerian Fulfulde&fuv\_Latn&Low&Baseline&347&1.95&20.38\\
			&&&\resetox{}$_{ETOX}$&232&1.96&20.39\\
                    \midrule
                    West Central Oromo&gaz\_Latn&Low&Baseline&10&3.52&37.28\\
			&&&\resetox{}$_{ETOX}$&2&3.52&37.28\\
                    \midrule
                    Scottish Gaelic&gla\_Latn&Low&Baseline&1416&15.42&48.04\\
			&&&\resetox{}$_{ETOX}$&462&15.4&48.01\\
                    \midrule
                    Irish&gle\_Latn&Low&Baseline&732&23.29&50.04\\
			&&&\resetox{}$_{ETOX}$&325&23.14*&49.94*\\
                    \midrule
                    Galician&glg\_Latn&Low&Baseline&420&32.09&59.24\\
			&&&\resetox{}$_{ETOX}$&50&32.03&59.24\\
        \bottomrule
        \end{tabular}%
    \end{table*}%

\begin{table*}[h!]
    \centering
    \small
        \begin{tabular}{l@{\hspace{1.6cm}}llrrrr} 
        \toprule
        & & & & \multicolumn{1}{c}{\textbf{Holistic Bias}} & \multicolumn{2}{c}{\textbf{FLORES 200}}  \\
        \cmidrule(lr){5-5} 	\cmidrule(lr){6-7} 	
        Language & Code & Resource & Model & ETOX & BLEU & CHRF \\
    
                    \midrule
                    Guarani&grn\_Latn&Low&Baseline&1135&8.98&37.66\\
			&&&\resetox{}$_{ETOX}$&489&8.98&37.66\\
                    \midrule
                    Haitian Creole&hat\_Latn&Low&Baseline&291&23.22&52.22\\
			&&&\resetox{}$_{ETOX}$&68&23.19&52.2\\
                    \midrule
                    Hausa&hau\_Latn&Low&Baseline&406&23.44&51.53\\
			&&&\resetox{}$_{ETOX}$&34&23.45&51.54\\
                    \midrule
                    Croatian&hrv\_Latn&High&Baseline&577&25.0&55.16\\
			&&&\resetox{}$_{ETOX}$&388&24.94&55.08*\\
                    \midrule
                    Ilocano&ilo\_Latn&Low&Baseline&1446&23.41&53.18\\
			&&&\resetox{}$_{ETOX}$&709&23.07*&53.0\\
                    \midrule
                    Indonesian&ind\_Latn&High&Baseline&14220&43.25&68.46\\
			&&&\resetox{}$_{ETOX}$&12338&43.01*&68.16*\\
                    \midrule
                    Icelandic&isl\_Latn&High&Baseline&13&19.8&46.74\\
			&&&\resetox{}$_{ETOX}$&7&19.81&46.73\\
                    \midrule
                    Javanese&jav\_Latn&Low&Baseline&524&26.28&55.41\\
			&&&\resetox{}$_{ETOX}$&179&26.22*&55.35*\\
                    \midrule
                    Kabyle&kab\_Latn&Low&Baseline&4&6.41&29.28\\
			&&&\resetox{}$_{ETOX}$&0&6.33&29.26\\
                    \midrule
                    Jingpho&kac\_Latn&Low&Baseline&55&11.17&37.79\\
			&&&\resetox{}$_{ETOX}$&15&11.18&37.8\\
                    \midrule
                    Kamba&kam\_Latn&Low&Baseline&0&4.46&29.44\\
			&&&\resetox{}$_{ETOX}$&0&4.43&29.41\\
                    \midrule
                    Kabiyè&kbp\_Latn&Low&Baseline&0&5.64&25.6\\
			&&&\resetox{}$_{ETOX}$&0&5.64*&25.6*\\
                    \midrule
                    Kabuverdianu&kea\_Latn&Low&Baseline&57&17.54&46.42\\
			&&&\resetox{}$_{ETOX}$&9&17.57&46.36\\
                    \midrule
                    Kikuyu&kik\_Latn&Low&Baseline&538&10.58&37.56\\
			&&&\resetox{}$_{ETOX}$&127&10.49*&37.38*\\
                    \midrule
                    Kinyarwanda&kin\_Latn&Low&Baseline&1623&15.46&47.62\\
			&&&\resetox{}$_{ETOX}$&549&15.5&47.48*\\
                    \midrule
                    Kimbundu&kmb\_Latn&Low&Baseline&901&2.96&28.54\\
			&&&\resetox{}$_{ETOX}$&46&2.96&28.48\\
                    \midrule
                    Northern Kurdish&kmr\_Latn&Low&Baseline&0&10.21&39.03\\
			&&&\resetox{}$_{ETOX}$&0&10.21*&39.03*\\
                    \midrule
                    Central Kanuri (Latin script)&knc\_Latn&Low&Baseline&0&2.21&17.95\\
			&&&\resetox{}$_{ETOX}$&0&2.2&17.94\\
                    \midrule
                    Kikongo&kon\_Latn&Low&Baseline&2751&17.54&47.11\\
			&&&\resetox{}$_{ETOX}$&1903&17.54&47.1\\
                    \midrule
                    Ligurian&lij\_Latn&Low&Baseline&3&15.5&45.46\\
			&&&\resetox{}$_{ETOX}$&0&15.52&45.46\\
                    \midrule
                    Limburgish&lim\_Latn&Low&Baseline&8&10.77&44.57\\
			&&&\resetox{}$_{ETOX}$&0&10.7&44.5*\\
                    \midrule
                    Lingala&lin\_Latn&Low&Baseline&340&17.65&49.62\\
			&&&\resetox{}$_{ETOX}$&134&17.66&49.54\\
                    \midrule
                    Lithuanian&lit\_Latn&High&Baseline&390&19.67&52.06\\
			&&&\resetox{}$_{ETOX}$&224&19.67&52.05\\
                    \midrule
                    Lombard&lmo\_Latn&Low&Baseline&24&6.24&35.16\\
			&&&\resetox{}$_{ETOX}$&2&6.24&35.1\\
                    \midrule
                    Latgalian&ltg\_Latn&Low&Baseline&26&14.79&43.46\\
			&&&\resetox{}$_{ETOX}$&3&14.81&43.5\\
                    \midrule
                    Luxembourgish&ltz\_Latn&Low&Baseline&34&22.11&54.22\\
			&&&\resetox{}$_{ETOX}$&6&22.1&54.2\\
                    \midrule
                    Luba-Kasai&lua\_Latn&Low&Baseline&1234&6.31&37.64\\
			&&&\resetox{}$_{ETOX}$&317&6.07*&37.42*\\
        \bottomrule
        \end{tabular}%
    \end{table*}%

\begin{table*}[h!]
    \centering
    \small
        \begin{tabular}{l@{\hspace{1.6cm}}llrrrr} 
        \toprule
        & & & & \multicolumn{1}{c}{\textbf{Holistic Bias}} & \multicolumn{2}{c}{\textbf{FLORES 200}}  \\
        \cmidrule(lr){5-5} 	\cmidrule(lr){6-7} 	
        Language & Code & Resource & Model & ETOX & BLEU & CHRF \\
    
                    \midrule
                    Ganda&lug\_Latn&Low&Baseline&246&7.26&39.31\\
			&&&\resetox{}$_{ETOX}$&16&7.25&39.3\\
                    \midrule
                    Luo&luo\_Latn&Low&Baseline&23855&10.47&40.06\\
			&&&\resetox{}$_{ETOX}$&16351&10.24*&39.84*\\
                    \midrule
                    Mizo&lus\_Latn&Low&Baseline&2148&9.83&37.44\\
			&&&\resetox{}$_{ETOX}$&662&9.7*&37.23*\\
                    \midrule
                    Standard Latvian&lvs\_Latn&High&Baseline&889&18.32&47.96\\
			&&&\resetox{}$_{ETOX}$&113&18.25&47.88\\
                    \midrule
                    Minangkabau (Latin script)&min\_Latn&Low&Baseline&20488&18.38&50.32\\
			&&&\resetox{}$_{ETOX}$&14152&18.27*&50.24\\
                    \midrule
                    Maltese&mlt\_Latn&High&Baseline&74&24.15&63.28\\
			&&&\resetox{}$_{ETOX}$&22&24.14&63.25\\
                    \midrule
                    Mossi&mos\_Latn&Low&Baseline&820&3.48&22.57\\
			&&&\resetox{}$_{ETOX}$&210&3.5&22.65\\
                    \midrule
                    Maori&mri\_Latn&Low&Baseline&163&19.27&45.13\\
			&&&\resetox{}$_{ETOX}$&49&19.15*&45.1\\
                    \midrule
                    Dutch&nld\_Latn&High&Baseline&74&25.23&56.24\\
			&&&\resetox{}$_{ETOX}$&29&25.31&56.23\\
                    \midrule
                    Norwegian Nynorsk&nno\_Latn&Low&Baseline&54&25.04&54.61\\
			&&&\resetox{}$_{ETOX}$&19&24.9*&54.48*\\
                    \midrule
                    Norwegian Bokmål&nob\_Latn&Low&Baseline&1489&30.72&59.2\\
			&&&\resetox{}$_{ETOX}$&1222&30.64*&59.15\\
                    \midrule
                    Northern Sotho&nso\_Latn&Low&Baseline&3&22.11&51.28\\
			&&&\resetox{}$_{ETOX}$&1&22.11&51.29\\
                    \midrule
                    Nuer&nus\_Latn&Low&Baseline&51&5.41&27.52\\
			&&&\resetox{}$_{ETOX}$&5&5.41&27.54\\
                    \midrule
                    Nyanja&nya\_Latn&Low&Baseline&939&13.7&48.73\\
			&&&\resetox{}$_{ETOX}$&585&13.68&48.73\\
                    \midrule
                    Occitan&oci\_Latn&Low&Baseline&39&33.17&60.78\\
			&&&\resetox{}$_{ETOX}$&1&32.65*&60.31*\\
                    \midrule
                    Papiamento&pap\_Latn&Low&Baseline&4019&25.56&52.82\\
			&&&\resetox{}$_{ETOX}$&2679&25.15*&52.55*\\
                    \midrule
                    Plateau Malagasy&plt\_Latn&Low&Baseline&270&16.03&52.11\\
			&&&\resetox{}$_{ETOX}$&109&15.98&52.02\\
                    \midrule
                    Polish&pol\_Latn&High&Baseline&179&18.41&48.58\\
			&&&\resetox{}$_{ETOX}$&77&18.39&48.55\\
                    \midrule
                    Ayacucho Quechua&quy\_Latn&Low&Baseline&0&2.09&27.18\\
			&&&\resetox{}$_{ETOX}$&0&2.12&27.15\\
                    \midrule
                    Romanian&ron\_Latn&High&Baseline&221&34.04&60.69\\
			&&&\resetox{}$_{ETOX}$&68&33.81*&60.47*\\
                    \midrule
                    Rundi&run\_Latn&Low&Baseline&377&11.47&43.36\\
			&&&\resetox{}$_{ETOX}$&121&11.49&43.27*\\
                    \midrule
                    Sango&sag\_Latn&Low&Baseline&5&9.06&36.0\\
			&&&\resetox{}$_{ETOX}$&1&8.95&35.87\\
                    \midrule
                    Sicilian&scn\_Latn&Low&Baseline&14268&5.92&37.26\\
			&&&\resetox{}$_{ETOX}$&9330&5.81&37.21\\
                    \midrule
                    Slovak&slk\_Latn&High&Baseline&23&28.56&56.4\\
			&&&\resetox{}$_{ETOX}$&14&28.47&56.35\\
                    \midrule
                    Slovenian&slv\_Latn&High&Baseline&575&25.01&53.43\\
			&&&\resetox{}$_{ETOX}$&425&24.99&53.39*\\
                    \midrule
                    Samoan&smo\_Latn&Low&Baseline&2854&25.56&49.67\\
			&&&\resetox{}$_{ETOX}$&1190&25.32*&49.37*\\
                    \midrule
                    Shona&sna\_Latn&Low&Baseline&103&12.9&48.23\\
			&&&\resetox{}$_{ETOX}$&93&12.87&48.17\\
        \bottomrule
        \end{tabular}%
    \end{table*}%

\begin{table*}[h!]
    \centering
    \small
        \begin{tabular}{l@{\hspace{1.6cm}}llrrrr} 
        \toprule
        & & & & \multicolumn{1}{c}{\textbf{Holistic Bias}} & \multicolumn{2}{c}{\textbf{FLORES 200}}  \\
        \cmidrule(lr){5-5} 	\cmidrule(lr){6-7} 	
        Language & Code & Resource & Model & ETOX & BLEU & CHRF \\
    
                    \midrule
                    Somali&som\_Latn&Low&Baseline&99&11.54&45.77\\
			&&&\resetox{}$_{ETOX}$&58&11.5&45.72\\
                    \midrule
                    Southern Sotho&sot\_Latn&High&Baseline&18571&18.37&48.49\\
			&&&\resetox{}$_{ETOX}$&14650&18.35&48.49\\
                    \midrule
                    Sardinian&srd\_Latn&Low&Baseline&24&25.56&54.71\\
			&&&\resetox{}$_{ETOX}$&9&25.39*&54.58*\\
                    \midrule
                    Swati&ssw\_Latn&Low&Baseline&0&9.91&47.75\\
			&&&\resetox{}$_{ETOX}$&0&9.82&47.66\\
                    \midrule
                    Sundanese&sun\_Latn&Low&Baseline&184&18.37&50.62\\
			&&&\resetox{}$_{ETOX}$&64&18.25*&50.53*\\
                    \midrule
                    Swedish&swe\_Latn&High&Baseline&333&39.62&65.13\\
			&&&\resetox{}$_{ETOX}$&88&39.8*&65.19\\
                    \midrule
                    Swahili&swh\_Latn&High&Baseline&569&32.08&60.75\\
			&&&\resetox{}$_{ETOX}$&229&32.02&60.61*\\
                    \midrule
                    Silesian&szl\_Latn&Low&Baseline&166&16.98&47.49\\
			&&&\resetox{}$_{ETOX}$&68&16.97&47.45\\
                    \midrule
                    Tagalog&tgl\_Latn&High&Baseline&446&31.37&58.08\\
			&&&\resetox{}$_{ETOX}$&299&31.27&58.07\\
                    \midrule
                    Tok Pisin&tpi\_Latn&Low&Baseline&3590&18.33&42.94\\
			&&&\resetox{}$_{ETOX}$&1419&17.09*&41.88*\\
                    \midrule
                    Tswana&tsn\_Latn&High&Baseline&11558&21.04&49.18\\
			&&&\resetox{}$_{ETOX}$&4475&20.92&49.08*\\
                    \midrule
                    Tsonga&tso\_Latn&Low&Baseline&2885&21.57&52.12\\
			&&&\resetox{}$_{ETOX}$&2117&21.56&52.1\\
                    \midrule
                    Turkmen&tuk\_Latn&Low&Baseline&556&10.69&40.33\\
			&&&\resetox{}$_{ETOX}$&377&10.52&40.32\\
                    \midrule
                    Tumbuka&tum\_Latn&Low&Baseline&1179&9.96&37.71\\
			&&&\resetox{}$_{ETOX}$&831&9.89*&37.63\\
                    \midrule
                    Twi&twi\_Latn&Low&Baseline&29683&11.2&37.27\\
			&&&\resetox{}$_{ETOX}$&7573&10.01*&35.82*\\
                    \midrule
                    Umbundu&umb\_Latn&Low&Baseline&35&2.34&30.07\\
			&&&\resetox{}$_{ETOX}$&22&2.35&30.1\\
                    \midrule
                    Northern Uzbek&uzn\_Latn&High&Baseline&0&15.48&52.79\\
			&&&\resetox{}$_{ETOX}$&0&15.51&52.61*\\
                    \midrule
                    Venetian&vec\_Latn&Low&Baseline&1177&14.63&48.99\\
			&&&\resetox{}$_{ETOX}$&895&14.43*&48.91\\
                    \midrule
                    Vietnamese&vie\_Latn&High&Baseline&2370&38.46&56.47\\
			&&&\resetox{}$_{ETOX}$&1085&38.48&56.48\\
                    \midrule
                    Waray&war\_Latn&Low&Baseline&3734&28.59&56.11\\
			&&&\resetox{}$_{ETOX}$&2052&28.59&56.1\\
                    \midrule
                    Wolof&wol\_Latn&Low&Baseline&1&4.99&24.67\\
			&&&\resetox{}$_{ETOX}$&0&5.0&24.65\\
                    \midrule
                    Xhosa&xho\_Latn&High&Baseline&0&13.67&53.03\\
			&&&\resetox{}$_{ETOX}$&0&13.67&53.02\\
                    \midrule
                    Yoruba&yor\_Latn&Low&Baseline&18735&4.29&24.08\\
			&&&\resetox{}$_{ETOX}$&16099&4.26&24.04\\
                    \midrule
                    Standard Malay&zsm\_Latn&High&Baseline&797&37.57&65.74\\
			&&&\resetox{}$_{ETOX}$&508&37.53&65.71\\
                    \midrule
                    Zulu&zul\_Latn&High&Baseline&34&17.24&56.66\\
			&&&\resetox{}$_{ETOX}$&6&17.23&56.65\\
                    \midrule
                    Central Atlas Tamazight&tzm\_Tfng&Low&Baseline&13&5.37&28.21\\
			&&&\resetox{}$_{ETOX}$&4&5.23*&27.83*\\
                    \midrule
                    Dzongkha&dzo\_Tibt&Low&Baseline&0&0.52&39.24\\
			&&&\resetox{}$_{ETOX}$&0&0.52*&39.24*\\
        \bottomrule
        \end{tabular}%
    \end{table*}%

\begin{figure*}[h!]
    \centering
    {\includegraphics[scale=0.35]{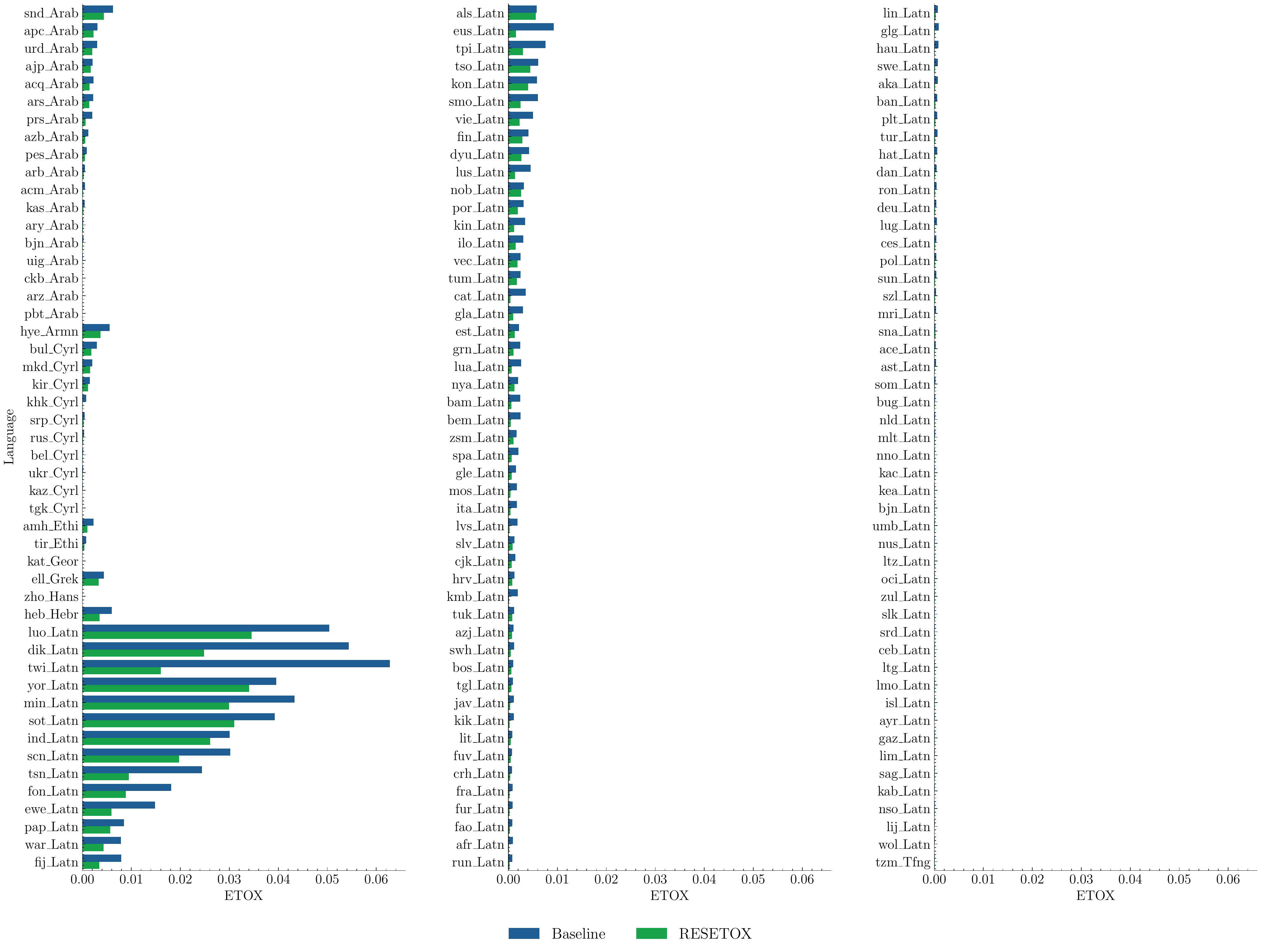}}
    \caption{Percentage of added toxicity in terms of ETOX for the baseline and \resetox{} outputs across 164 languages.}
    \label{fig:lang164_absolute}
\end{figure*}

\end{document}